\documentclass[12pt]{article}
\usepackage[utf8]{inputenc}
\usepackage[letterpaper, margin=1in]{geometry}

\usepackage[font={small}]{caption}
\usepackage{subcaption}
\usepackage{graphicx}
\usepackage{stackengine}
\usepackage{appendix}
\usepackage{rotating}
\usepackage{float}
\usepackage[sorting=none]{biblatex}
\usepackage{csquotes}

\usepackage{bm}
\usepackage{amsmath}
\usepackage{amssymb}

\usepackage[hidelinks]{hyperref}
\usepackage{cleveref}

\DeclareMathOperator*{\argmin}{arg\,min}
\newcommand{\vect}[1]{\ensuremath{\boldsymbol{ #1 }}}

\newcommand{\todocite}[1]{\textbf{\{ CITE \}}}

\graphicspath{ {images/} }

\title{Exploring Complex Dynamical Systems via Nonconvex Optimization}
\author{Hunter Elliott}
\date{December 2022}

\begin{document}

\maketitle

\begin{abstract}
    Cataloging the complex behaviors of dynamical systems can be challenging, even when they are well-described by a simple mechanistic model. If such a system is of limited analytical tractability, brute force simulation is often the only resort. We present an alternative, optimization-driven approach using tools from machine learning. We apply this approach to a novel, fully-optimizable, reaction-diffusion model which incorporates complex chemical reaction networks (termed ``Dense Reaction-Diffusion Network'' or ``Dense RDN''). This allows us to systematically identify new states and behaviors, including pattern formation, dissipation-maximizing nonequilibrium states, and replication-like dynamical structures.
\end{abstract}

\section{Introduction}
\newcommand{\exampanelscale}{0.1\textwidth}

\begin{figure}[b]
    \centering
    \begin{subfigure}[b]{\exampanelscale}
        \includegraphics[width=\textwidth]{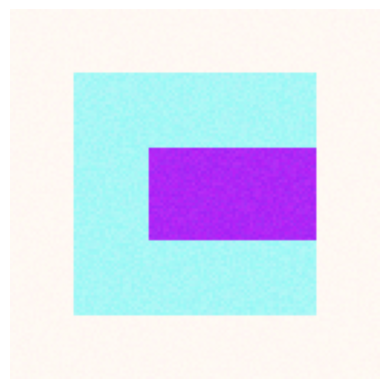}
    \end{subfigure}
    \begin{subfigure}[b]{\exampanelscale}
        \includegraphics[width=\textwidth]{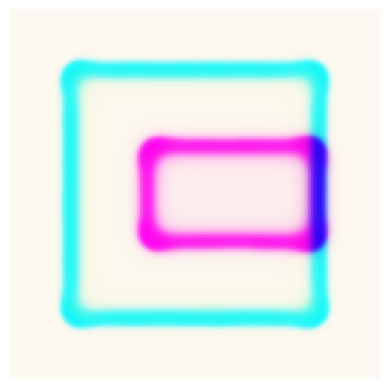}
    \end{subfigure}
    \begin{subfigure}[b]{\exampanelscale}
        \includegraphics[width=\textwidth]{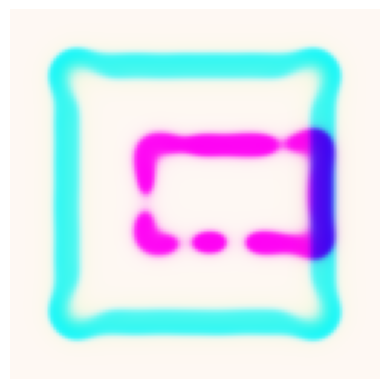}
    \end{subfigure}
    \begin{subfigure}[b]{\exampanelscale}
        \includegraphics[width=\textwidth]{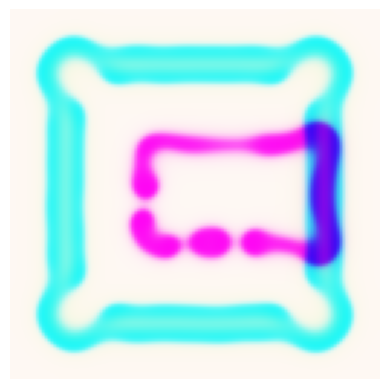}
    \end{subfigure}
    \begin{subfigure}[b]{\exampanelscale}
        \includegraphics[width=\textwidth]{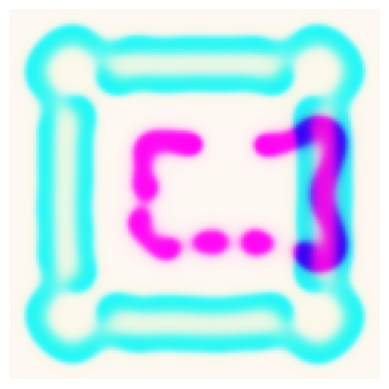}
    \end{subfigure}
    \begin{subfigure}[b]{\exampanelscale}
        \includegraphics[width=\textwidth]{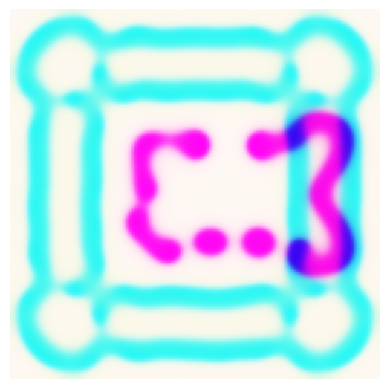}
    \end{subfigure}
    \begin{subfigure}[b]{\exampanelscale}
        \includegraphics[width=\textwidth]{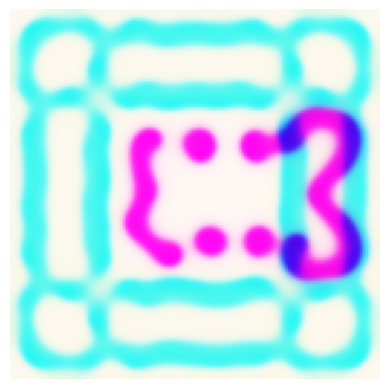}
    \end{subfigure}
    \begin{subfigure}[b]{\exampanelscale}
        \includegraphics[width=\textwidth]{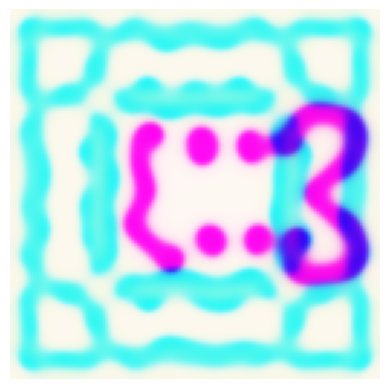}
    \end{subfigure}
    \begin{subfigure}[b]{\exampanelscale}
        \includegraphics[width=\textwidth]{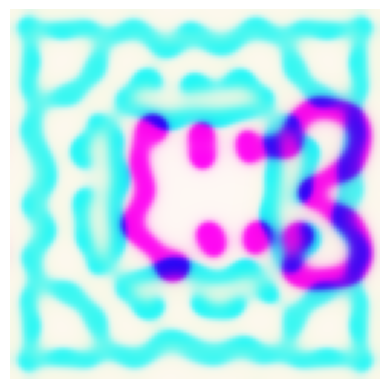}
    \end{subfigure}
    
    \caption{An example of the complex dynamics possible in reaction-diffusion systems. Each panel shows the local concentration of 5 chemical species, from $t=0s$ (left) to $t=1000s$ (right). All concentration visualizations are PCA projected to 3 colors and in arbitrary units (Appendix \ref{sec:conc_vis}).}
    \label{fig:complex_example}
\end{figure}

Chemical reaction systems driven far from equilibrium can demonstrate striking complexity in both the temporal and spatial variations of the concentration of their constituent chemical species (Figure \ref{fig:complex_example}, Appendix \ref{sec:complex_example_description}, \cite{pearson1993complex}). This complexity can be captured in simple reaction-diffusion (RD) models, and has been appreciated to be relevant for understanding a variety of nonequilibrium phenomena ranging from biological pattern formation \cite{kondo2010reaction}\cite{landge2020pattern} to the emergence of entropy-producing “dissipative structures” more broadly \cite{prigogine1968symmetry}, and has even been speculated to be important for the earliest stages of abiogenesis \cite{prigogine1971biological}\cite{szabo2002silico}\cite{adamski2020self}. The underlying microscopic physicochemical principles are well understood, and with minimal assumptions simple differential models incorporating these principles agree well with experiment \cite{lee1993pattern}\cite{armstrong2004modelling}\cite{ertl1991oscillatory}\cite{hamik2003excitation}.

Despite this mechanistic understanding, a reductionist approach to examining these systems often bears little fruit. Writing down the partial differential equations describing the system does not easily yield a full picture of the allowed behaviors via analytical means, requiring instead a focus on reduced models or specific behaviors \cite{or1998spot}\cite{khater2002tanh}\cite{smith2018beyond}\cite{kondo2017updated}. Some analytical approaches may define parameter ranges corresponding to stable or complex dynamics, but cannot in the general case enumerate \textit{which} complex behaviors and states are attainable \cite{feinberg2019foundations}. This is further complicated by the fact that many of these systems display chaotic behavior, implying that over time they may visit an infinity of states \cite{volpert2009reaction}\cite{simoyi1982one}\cite{rossler1976chemical}. Forward numerical simulation can be carried out with high accuracy, yet for systems of even moderate complexity a brute-force exploration of state and parameter space is at best tedious and at worst prohibitively computationally intensive \cite{landge2020pattern}.

One alternative ``inverse desig'' approach, explored here, is to first choose a specific behavior of interest and then ask whether points in the configuration space of a model can be found which correspond to this behavior. If the behavior of interest can be formulated as an easily evaluable mathematical function of the system state and or dynamics, this allows the exploration of the system’s behavior to be framed as an optimization problem. While this optimization problem is in most cases nonconvex, some tools from modern machine learning (ML) may still be applied. In this way rather than attempting to fully catalog the possible behaviors a chemical reaction system can display, we instead test a hypothesis regarding a particular state or dynamic of interest, side-stepping both the analytical intractability and the prohibitive complexity of brute force search.

The inverse design of RD systems has previously been framed as an optimization problem, albeit without physically realistic chemical models and with applications often aimed at image processing or texture synthesis \cite{phan2006sketching}\cite{bartocci2016formal}\cite{mordvintsev2021differentiable}\cite{mordvintsev2021texture}. Some have designed realistic RD systems, but using heuristics \cite{tikhomirov2017fractal}\cites{scholes2017three}. Other inverse design approaches working with physically realistic RD models avoid optimization and instead aim to produce RD system modules that compartmentalize complexity and allow programmatic design of modular or hierarchical structures \cite{scalise2014designing} or cellular automata-like boolean dynamics \cite{scalise2016emulating}. Some methods allow inverse design of CRN dynamics but without a diffusion component or spatial organization \cite{murphy2018synthesizing}. Machine learning approaches have been applied to related realistic models, but in unrelated ways; either as methods to approximate the forward model \cite{chua1995autonomous}\cite{li2020reaction}\cite{pathak2018model}, or to learn models from real world data \cite{pathak2017using}\cite{zakeri2019weakly}\cite{yeo2019deep}. Here we do not approximate the physical model nor do we use any real world data. Instead we use only the optimization approaches from ML to explore not just steady states but dynamics and thermodynamics of an un-approximated, known, yet flexible, physically realistic forward model.

The approach we present has it’s limitations, not only in the systems and behaviors which can be explored, but also in the conclusions that can be formed from these explorations: The failure of the optimization to converge is an absence of proof that a state or dynamic is possible, not a proof of absence. Still, we aim to demonstrate here that it can complement existing approaches and may find use in investigations not only of driven nonequilibrium chemical reaction systems but in complex dynamical systems more broadly.

\section{An Optimization Approach}

At a high level, this approach requires that we specify a model which simulates the dynamics of the system of interest, as well as a loss function representing a hypothesis about a dynamical behavior the system \textit{may} be capable of. The loss is simply a scalar function of the model's state, dynamics, and parameters which takes on small values if and only if the behavior or state of interest is exhibited. If the model and loss are both differentiable, we can use model construction and optimization approaches from machine learning to minimize the loss and attempt to realize the behavior of interest. Here we focus on a novel 'dense reaction-diffusion network' model or ``Dense RDN'', where an arbitrary number of chemical species interchange via a reaction network while also diffusing in two spatial dimensions. The chemical reactions, their rate constants, the diffusion coefficients, and the initial conditions are all determined by the optimization. The loss function is of the investigator's choice, and could represent simple behaviors like stability or bi-stability, or more complex behaviors, of which we present several examples in Section \ref{sec:results}.

\subsection{Preliminaries}
More formally, we assume that it is possible to specify a forward model $\Psi(\bm{X},\vect{\theta}_\Psi)$ which propagates a state $\bm{X}$ through time, parameterized by $\vect{\theta}_\Psi$. The model should give differentials $\frac{\partial \bm{X}}{\partial t}=\Psi(\bm{X},\vect{\theta}_\Psi)$ which can be used in \textit{e.g.} forward Euler integration:

\begin{equation}\label{eq:euler_int}
  \begin{aligned}
    \bm{X}_{t+1} &= \bm{X}_t + \Psi(\bm{X}_t,\vect{\theta}_\Psi) \Delta t \\
                 &= \bm{X}_t + \Delta \bm{X}_{t} 
  \end{aligned}
\end{equation}
With a time step size $\Delta t$ giving a change in state $\Delta \bm{X}_{t}$ (see Appendix \ref{sec:step_size} for details on how we ensure this step size is appropriately small).
The initial conditions $\bm{X}_0$ are generated from a random vector $\vect{z}$ by a model $\bm{X}_0=G(\vect{z},\vect{\theta}_G)$ (in this work a neural network) parameterized by $\vect{\theta}_G$. Repeated application of (\ref{eq:euler_int}) to these initial conditions gives a time evolution $\mathcal{X}=\left[ \bm{X}_0, \bm{X}_1,...,\bm{X}_T\right]$. We define  $\vect{\theta}=\vect{\theta}_\Psi \cup \vect{\theta}_G$ for convenience and require that both $\Psi$ and $G$ be differentiable almost everywhere with respect to both $\bm{X}$ and $\vect{\theta}$.

Finally, we assume a scalar loss function $\mathcal{L}(\vect{\theta})$ can be specified such that as the loss approaches a minimum value the behavior of interest is exhibited in $\mathcal{X}$. As a simple example, if we sought to identify steady states we could define $\mathcal{L}(\vect{\theta}) = \lVert \Delta \bm{X}_t \rVert_2$. Thus, in general we seek parameters $\vect{\theta}^*$:

\begin{equation}
    \vect{\theta}^* = \argmin_{\vect{\theta}} \mathcal{L}(\vect{\theta})    
\end{equation}

Because $\mathcal{L}$ is fully differentiable, gradient-based optimization techniques such as stochastic gradient descent (SGD) can be applied. If the optimization converges to a sufficiently low value of the loss such that the resulting dynamics meet the investigator's criteria, then we will have determined initial conditions and transition model parameters which result in a time evolution $\mathcal{X}$ that exhibits the behavior of interest, thus proving it is within the possible dynamics of the model. In contrast, if the optimization fails to yield such parameters, we cannot conclude that the desired behavior is impossible, only that this procedure failed to parameterize it.

\paragraph{An expectation of success?}
At first glance this may seem a doomed endeavor. Our expressed interest is in exploring models $\Psi(\bm{X},\vect{\theta})$ with intractably complex, nonlinear dynamics. This makes the relationship between $\vect{\theta}$ and $\mathcal{X}$ highly nonlinear and provides no guarantee of convexity in our loss $\mathcal{L}$. Nonconvex optimization is difficult (NP-hard), yet we propose to use local optimization methods such as SGD to find $\vect{\theta}^*$. Nonetheless we are encouraged by two observations. First, we do not require that we find a global minimum of the loss, only a 'sufficiently low' local minimum, such that the resulting structure and/or dynamics meet the investigator's criteria for the behavior of interest. Second, the entire field of deep learning (DL, a sub-field of machine learning), a field which has seen remarkable success, relies on such optimizations succeeding despite their apparent in-feasibility. The reasons for the empirically observed reliable convergence of such nonconvex optimizations is still an active area of research, and not addressed here. However, one proposed explanation which we conjecture is of relevance here is this: When the parameter space being optimized is of sufficiently high dimensionality, the existence of true local minima with high loss values becomes increasingly unlikely \cite{dauphin2014identifying}. We do not systematically study the conditions required for convergence.  We do however demonstrate that by embedding high-dimensional physicochemical models within yet higher dimensional neural networks and employing optimization approaches used in DL, we are able to achieve convergence with a variety of loss functions. This at least demonstrates empirically that the optimization approach popularized in data driven machine learning can be applied to a data-free, purely forward simulation-driven exploration of complex dynamical models.

\subsection{A Dense Reaction-Diffusion Network Model}

We chose reaction-diffusion (RD) models as a simple yet accurate model which can display spatiotemporally complex behaviors and which has at least conceptual relevance for physics, chemistry, and biology. Even simple RD systems have been observed experimentally to produce complex and even chaotic behavior which is well described by these models \cite{lee1993pattern}\cite{armstrong2004modelling}\cite{ertl1991oscillatory}\cite{hamik2003excitation}. When driven by the constant influx of reactants, these models represent nonequilibrium systems of interest in thermodynamics and possibly even abiogenesis \cite{prigogine1971biological}\cite{scheuring2003spatial}. While much prior work has analyzed hand-designed chemical reaction sets, here we begin with a large, dense network of chemical reactions and allow the appropriate reaction set to be determined during the optimization.

\subsubsection{Chemistry: Chemical Reaction Network}
In this work the forward model $\Psi$ is a reaction-diffusion model. The chemical reactions in this model comprise what we refer to as a ``dense'' chemical reaction network (CRN), because it contains every possible reaction which matches these three reaction prototypes:

\begin{equation}
    \begin{aligned}
        A + B &\rightleftarrows 2C \\  \label{eq:crn_sketch}
        A &\rightleftarrows B \\   
        A + 2B &\rightleftarrows 3B \\   
    \end{aligned}
\end{equation}

By ``reaction prototype'' we mean that, while the CRN may contain any number of chemical species $N_s$, the letters in \eqref{eq:crn_sketch} simply indicate that $A$,$B$,$C$ must be 3 \textit{different} species, with the specified stoichiometries (See Appendix \ref{sec:crn_methods}).
The forward and reverse rates of each reaction are free parameters $\mathcal{\bm{\kappa}}=\{k_{f1}, k_{r1}, k_{f2}, k_{r2},...\}\subset\vect{\theta}_\Psi$ which, with standard mass action kinetics, yield a net change in state due to chemical reactions $\Delta\bm{X}_{t_{Rxn}}$. With the reaction types in (\ref{eq:crn_sketch}), the resulting dynamics are highly nonlinear, and we can be assured that complex behavior is at least \textit{possible} (See Appendix \ref{sec:crn_motivation}), despite much of the kinetic parameter space producing uninteresting behavior (\textit{e.g.} monotonic relaxation to equilibrium). This reaction network structure also encompasses much of the uni-, bi-, and tri-molecular reactions possible amongst 3 or more chemical species.

While the simultaneous existence of all of these reactions is perhaps not probable in a real-world CRN, note that the optimization can set reaction rates to $\sim0$, and so we are effectively optimizing not only for the rates of reactions but also which reactions to include.

\subsubsection{Nonequilibrium: Flow Reactor Drive}

We model the system as being within a 'flow reactor', so it is maintained away from equilibrium by a constant influx of reactants. This influx produces a corresponding outflow, giving a net change in concentration for chemical species $\bm{X}^i$ due to this drive of \cite{kondepudi2014modern}\cite{pearson1993complex}:

\begin{equation}
    \Delta \bm{X}_{t_{Drv}}^i = \left(f x_i - f\bm{X}_t^i \right) \Delta t
\end{equation}

Where again both the per-species feed concentrations $x_i$ and the shared flow rate $f$ are determined during the optimization. 

\subsubsection{Space: Diffusion and Initial Conditions}

To allow for spatial organization, the CRN exists within a discretized two dimensional domain such that the concentration of chemical species $i$ at position $(u, v)$ is given by $\bm{X}^i(u, v)$. The initial conditions $\bm{X_0}$ are generated from random vector $\vect{z}$ by $G$ which is instantiated as a neural network, similar to convolutional generator models such as DCGAN \cite{radford2015unsupervised} (see Appendix \ref{sec:dnns} for details).

Each chemical species also undergoes diffusion:

\begin{equation}\label{eq:diff_update}
    \Delta \bm{X}_{t_{Dif}}^i = D_i \nabla^2\bm{X}_t^i \Delta t
\end{equation}

With an optimizable diffusion coefficient $D_i \subset \vect{\theta}_\Psi$. The final combined change in state then is the sum of the contributions from reactions, drive, and diffusion:

\begin{equation}
    \Delta \bm{X}_t = \Delta\bm{X}_{t_{Rxn}} + \Delta\bm{X}_{t_{Drv}} + \Delta\bm{X}_{t_{Dif}}
\end{equation}

Together these terms and their parameters define a space of nonequilibrium physicochemical models which is capable of both mundane and spatiotemporally complex behavior, and is flexible enough to allow the optimization procedure to determine the actual states and dynamics it adopts.

\section{Results}\label{sec:results}
All that remains to fully specify an optimization is the loss function $\mathcal{L}$ encoding a behavior of interest. In effect, this loss function encodes an hypothesis about the possible states and/or dynamics of the model, and we investigate several such hypotheses here.

\subsection{Pattern Formation}\label{sec:pattern_formation}

Pattern formation - emergent structured spatial variation in the concentrations of chemical species - is a hallmark behavior of reaction-diffusion systems \cite{kondo2010reaction}\cite{vanag2009pattern}. Since postulated by Turing in 1952 \cite{turing1952chemical} it has been repeatedly recapitulated experimentally and is understood to be important for biological pattern formation \cite{kondo2010reaction}\cite{landge2020pattern} and even hypothesized to be relevant for the earliest stages of abiogenesis \cite{prigogine1971biological}\cite{szabo2002silico}\cite{adamski2020self}. 

\begin{figure}[hb]
    \centering
        \includegraphics[width=\textwidth]{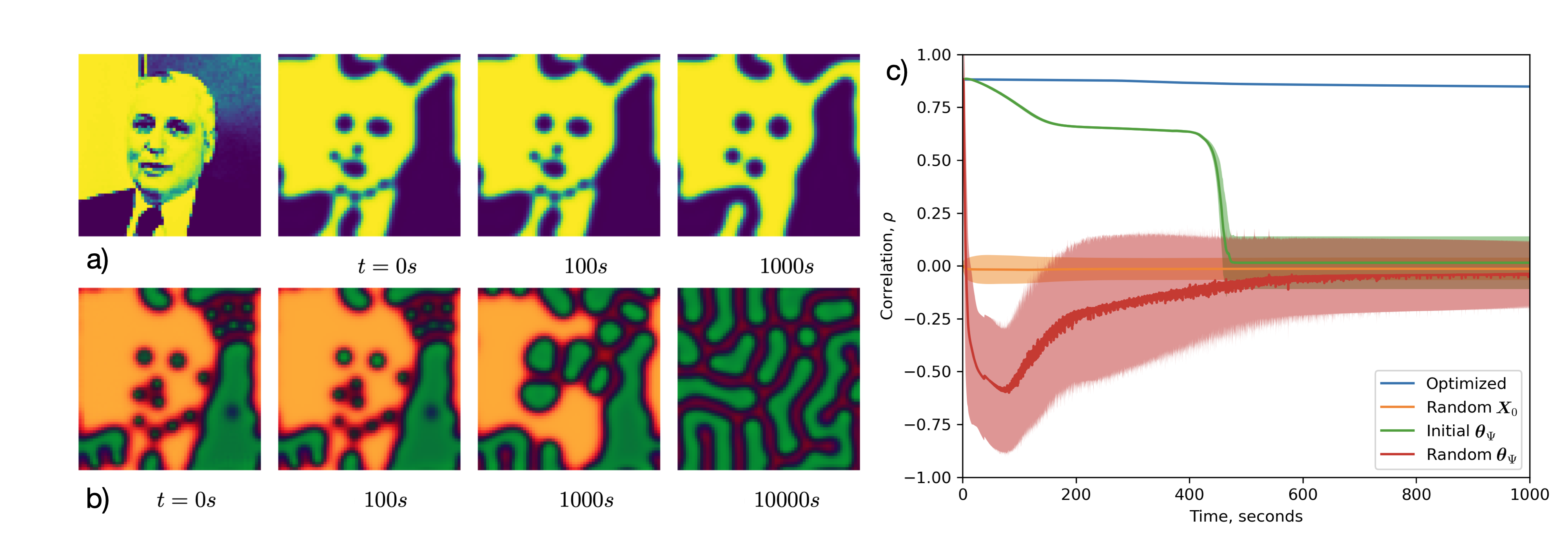}    
    \caption{A 4-chemical species dense reaction-diffusion network optimized to match a target (a, leftmost panel) supports semi-stable dissipative structures (a, right panels). A second 5-chemical species example (b) is less stable, showing dynamic pattern formation. c) Measuring correlation over time for the example in a) confirms that the optimization of both the initial conditions and reaction network are necessary for fidelity and stability. Solid lines show means, bands show +/- one standard deviation over $n=32$ randomizations.}
    \label{fig:fixed_ilya}
\end{figure}

We seek to identify kinetic parameter ($\vect{\theta}_\Psi$) regimes which correspond to the ability of the system to support stable patterns of arbitrary structure. As these will be driven nonequilibrium patterns, we choose a fitting arbitrary structure as our 'target': An Image of Ilya Prigogine (Figure \ref{fig:fixed_ilya}a, left) as a tribute to his seminal work on dissipative structures \cite{prigogine1968symmetry}\cite{prigogine1978time}. We therefore define a loss term which encourages the concentration distribution of chemical species $i$ to match this target, $\tilde{\bm{X}}$:

\begin{equation}\label{eq:target_loss}
    \mathcal{L}(\vect{\theta}) = \mathbb{E} \left[ \lVert \Psi(\bm{X}_{t-1}, \vect{\theta}_\Psi)^i - \tilde{\bm{X}} \rVert_2 \right] = \mathbb{E} \left[ \lVert \bm{X}_{t}^i - \tilde{\bm{X}} \rVert_2 \right]
\end{equation}

Where the expectation is taken both over time and $\bm{X}_0 \sim G(\vect{z}, \vect{\theta}_G)$, in practice implemented via random sampling in time and from \vect{z}. Furthermore we aim to identify \textit{stable} patterns, so we introduce an additional loss term which encourages temporal stability:

\begin{equation}\label{eq:target_stable_loss}
    \mathcal{L}(\vect{\theta}) = \mathbb{E} \left[ \lVert \bm{X}_{t}^i - \tilde{\bm{X}} \rVert_2 + \lambda_\Delta\lVert \Delta \bm{X}_t \rVert_1 \right]
\end{equation}

This ensures that we are not just optimizing initial conditions which match the target, but also a dynamical model which preserves it over time. Successful convergence to patterns which mimic this target requires many of the optimization heuristics that have become commonplace in deep learning, as well as some task- and model-specific adjustments, detailed in Appendix \ref{sec:optimization}.
Nonetheless it is possible, as shown in Figure \ref{fig:fixed_ilya}a, where a pattern found in a 4-species CRN shows reasonable fidelity to the target (Average Pearson's $r$ of .88 over the optimized timescale of $32s$), as well as stability over timescales dramatically longer than those employed during the optimization (Figure \ref{fig:fixed_ilya}c). In other experiments the target pattern is not as temporally stable, and the optimization has clearly converged to a kinetic regime corresponding to dynamic pattern formation (Figure \ref{fig:fixed_ilya}b). We confirm that the optimization of both the initial conditions $\bm{X}_0$ as well as the reaction network parameters $\vect{\theta}_\Psi$ are necessary, by randomizing these and comparing the resulting correlation with the target over time (Figure \ref{fig:fixed_ilya}c, see also Appendix \ref{sec:controls}). 

This therefore confirms that we can \textit{de novo} identify a combination of initial conditions and reaction network kinetics and topologies (See Appendix \ref{sec:crn_graphs}) which support the formation of arbitrary patterns purely through optimization.

\subsection{Dissipation Maximization}\label{sec:dissipation_maximization}

As these are driven nonequilibrium systems it is natural to ask how the behavior of these models varies as the rate of entropy production - the dissipation rate - increases. In a system which is not relaxing towards equilibrium the dissipation rate is the rate at which input drive energy is converted to entropy. The thermodynamics of this Dense RDN system are well defined(\cite{kondepudi2014modern}\cite{mahara2005entropy}, see Appendix \ref{sec:therm}). The entropy production rates are differentiable functions of the time evolution $\mathcal{X}$, and so can be used to formulate a loss function, broken down in terms of the diffusion and reaction dissipation rates $\sigma_{Tot} = \sigma_{Rxn} + \sigma_{Dif}$:

\begin{equation}\label{eq:base_entropy_loss}
    \mathcal{L}(\vect{\theta}) = \mathbb{E} \left[ e^{-\sigma_{Rxn}(\bm{X}_t,\vect{\theta}) - \sigma_{Dif}(\bm{X}_t,\vect{\theta})} \right]
\end{equation}
\begin{figure}[!b]
    \centering
    \includegraphics[width=\textwidth]{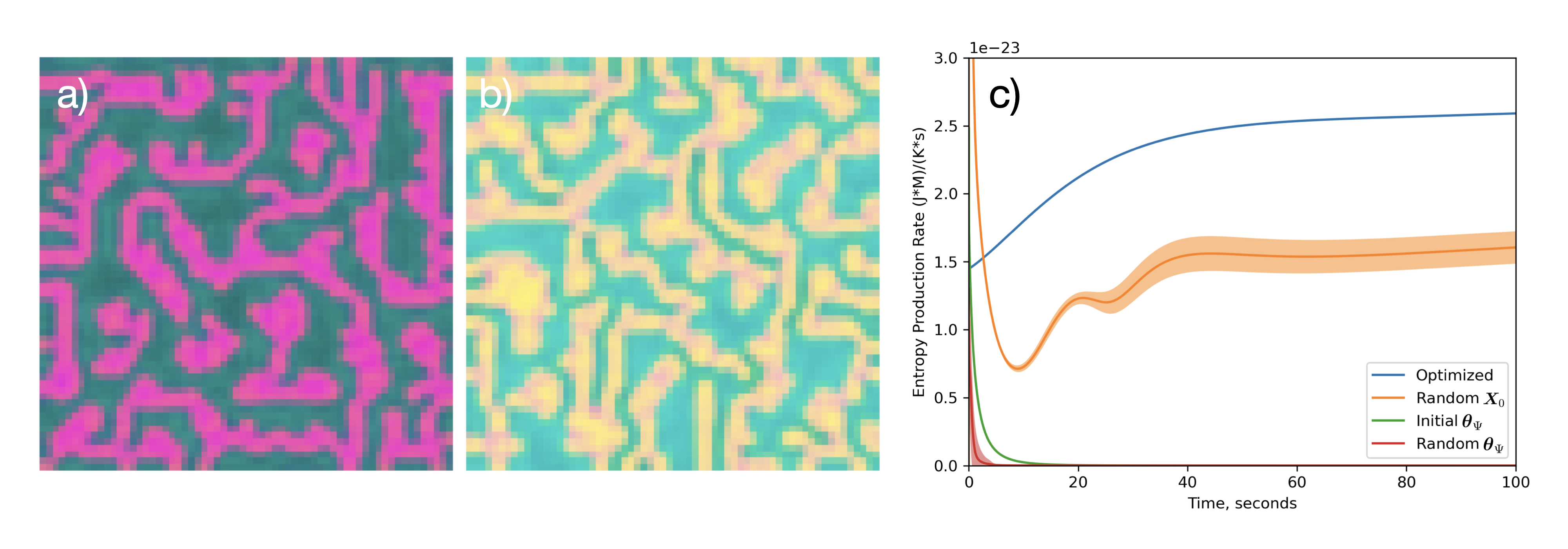}
    \caption{Maximizing the rate of entropy production from diffusion gives interesting filamentous structures (a-b) in the local concentrations of a 5-chemical species Dense RDN. We confirm the contribution of each optimized component in (c) by a comparison of the  dissipation rate vs. time for the optimized example in a) (blue) to those with randomized initial conditions (orange), the parameters the optimization was initialized to (green) and models with randomized kinetics (red). Solid lines show means, bands show +/- one standard deviation over $n=32$ randomizations.}
     \label{fig:diss_max}
\end{figure}
Note that we exponentiate the negative dissipation rates so that this is a minimization problem bounded below by zero; the raw dissipation rates are not bounded and can vary by several orders of magnitude, which can induce numerical instability in the optimization. Finally, we seek states which are \textit{stably} nonequilibrium, rather than those which are simply undergoing dissipative relaxation to equilibrium, so we introduce a term which encourages solutions for which the dissipation rate is unchanging:

\begin{equation}\label{eq:stable_entropy_loss}
    \mathcal{L}(\vect{\theta}) = \mathbb{E} \left[ e^{-\sigma_{Rxn}(\bm{X}_t,\vect{\theta)}) - \sigma_{Dif}(\bm{X}_t,\vect{\theta)})} \right] + \lambda_\sigma \mathrm{Var}\left[e^{-\sigma_{Tot}}\right]
\end{equation}

Where the variance is only over time and where the drive flow rate $f$ is constrained so that the maximum dissipation rate is finite. Minimization of (\ref{eq:stable_entropy_loss}) is dominated by the reaction dissipation rate, which is invariant under permutation of spatial positions and so unsurprisingly does not induce any spatial structure. Still, we can examine the resulting reaction networks as examples of high-entropy generation rate CRNs (see Appendix \ref{sec:crn_graphs} for a visualization of the CRN. Note in this example the drive flow rate $f$ is fixed to $.03s^{-1}$). 

If we instead include only the diffusion dissipation rate term, which \textit{requires} spatial non-uniformity in concentration to be non-zero, we find interesting structures which stably dissipate drive energy at a rate well above those seen without optimization and over timescales much longer than the 64 seconds used during optimization (Figure \ref{fig:diss_max}). It is worth noting that the form of these structures was not in any way pre-specified but rather emerges purely from the interaction between the dissipation-maximization loss function and the physicochemical properties of the model.

\subsubsection{Dissipative Distributions}\label{sec:inf_transmit}

In practice all of the preceding optimizations suffer from 'mode collapse' in the generative model $G$ such that it converges to producing only a single initial condition (meaning that the expectations in (\ref{eq:target_stable_loss}) (\ref{eq:stable_entropy_loss}) are effectively over time only). It may be desirable to instead produce a distribution, not only of $\bm{X}_0$ but of its time evolution as well. We can enforce this by introducing a decoder model $\hat{\vect{z}}_t=E(\bm{X}_t)$ which is optimized to reconstruct the random binary vector $\vect{z}$ used to generate $\bm{X}_0$, via an associated loss term:
\begin{equation}\label{eq:z_recon_term}
    \mathcal{L}_z(\vect{\theta},\hat{\vect{z}})=\mathbb{E} \left[ H(\vect{z},\hat{\vect{z}}_t) \right]
\end{equation}
Where $H$ is the cross entropy and $E$ is implemented as a convolutional neural network. With a deterministic forward model \eqref{eq:z_recon_term} is trivially satisfiable, so we also introduce a spatiotemporally variable, uniformly distributed noise term $\bm{\epsilon}_f(u,v)$ which corresponds to stochastic variation in the flow rate:
\begin{equation}
    \Delta \bm{X}_{t_{Drv}}^i = (f + \bm{\epsilon}_f)\odot\left(x_i - \bm{X}_t^i \right) \Delta t
\end{equation}
With $\bm{\epsilon}_f(u,v) \sim \mathrm{Unif}(0,0.8)$ (Gaussian low-pass filtered to avoid spatial discontinuities) and $\odot$ indicating the Hadamard product. This serves to ensure some approximate $\epsilon_f$-ball separation between the samples of $\bm{X}_t$.

\newcommand{\distpanelscale}{0.12\textwidth}

\begin{figure}[h]
    \centering
    \begin{subfigure}[b]{\distpanelscale}
        \includegraphics[width=\textwidth]{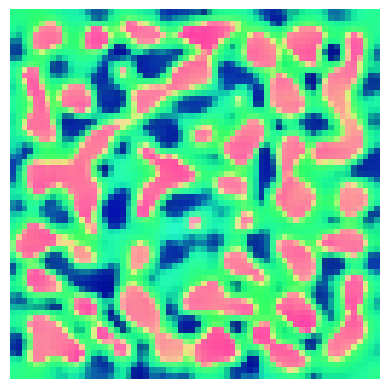}
    \end{subfigure}
    \begin{subfigure}[b]{\distpanelscale}
        \includegraphics[width=\textwidth]{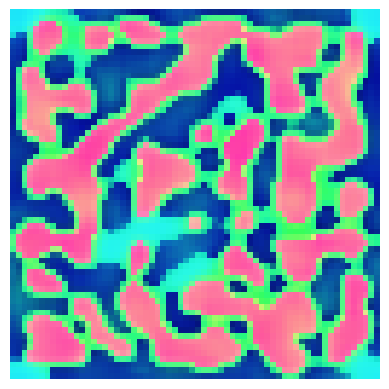}
    \end{subfigure}
    \begin{subfigure}[b]{\distpanelscale}
        \includegraphics[width=\textwidth]{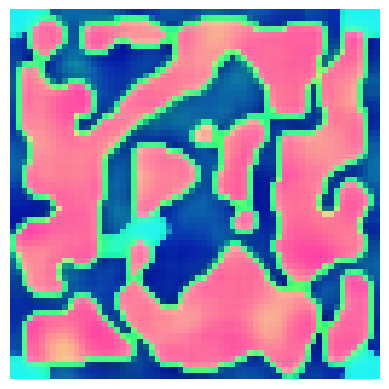}
    \end{subfigure}
    
    \begin{subfigure}[b]{\distpanelscale}
        \includegraphics[width=\textwidth]{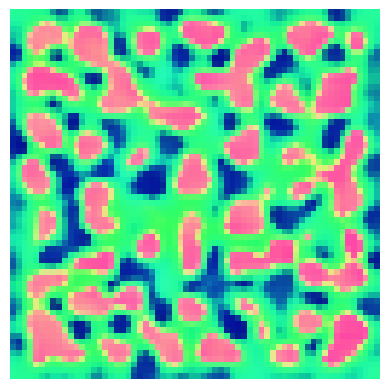}
    \end{subfigure}
    \begin{subfigure}[b]{\distpanelscale}
        \includegraphics[width=\textwidth]{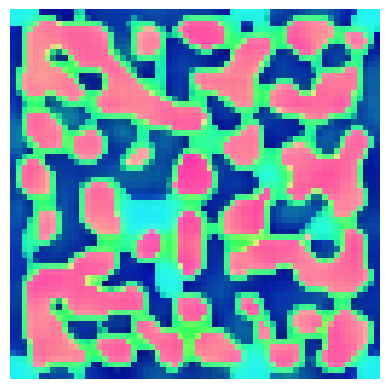}
    \end{subfigure}
    \begin{subfigure}[b]{\distpanelscale}
        \includegraphics[width=\textwidth]{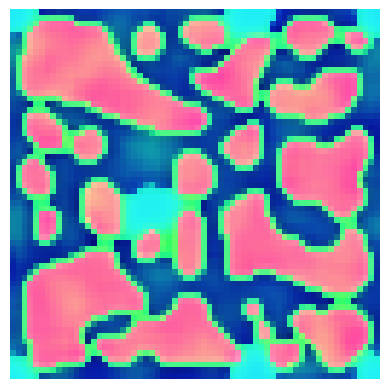}
    \end{subfigure}
    
    \begin{subfigure}[b]{\distpanelscale}
        \includegraphics[width=\textwidth]{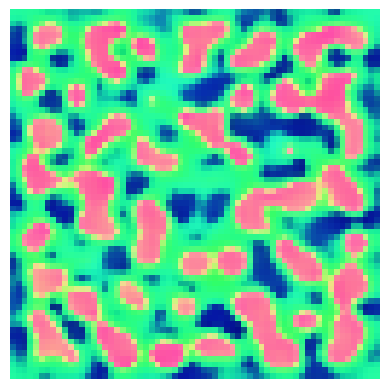}
    \end{subfigure}
    \begin{subfigure}[b]{\distpanelscale}
        \includegraphics[width=\textwidth]{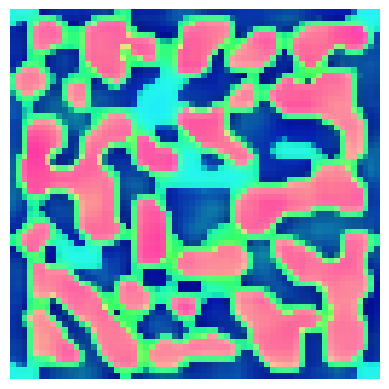}
    \end{subfigure}
    \begin{subfigure}[b]{\distpanelscale}
        \includegraphics[width=\textwidth]{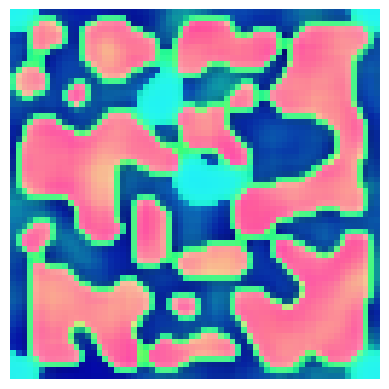}
    \end{subfigure}
    \caption{Distributions of dissipative structures in a 5-chemical species Dense RDN found by simultaneously maximizing diffusive entropy production and a loss that requires the $\vect{z}$ vector used to generate the initial conditions be reconstructable at every time point. Each row is a sample from $\vect{z}$ and the columns correspond to $t=10, 100, 1000s$ from left to right.}
    \label{fig:diss_dist}
\end{figure}

Minimizing a the sum of (\ref{eq:stable_entropy_loss}) and (\ref{eq:z_recon_term}) yields $\vect{\theta^*}$ corresponding to a distribution of states which maintain their uniqueness over time, despite the randomly fluctuating drive (Figure \ref{fig:diss_dist}). In the example shown the decoder is able to reconstruct the 16-bit $\vect{z}$ vector with 100\% accuracy for over 1000 time points, thus implying that this dynamical system is capable of transmitting information through time with a channel capacity proportional to the Shannon entropy of $\vect{z}$, despite the noisy environment.

\subsection{Dynamic Structure}\label{sec:replication}

\begin{figure}
    \centering
    \includegraphics[width=\textwidth]{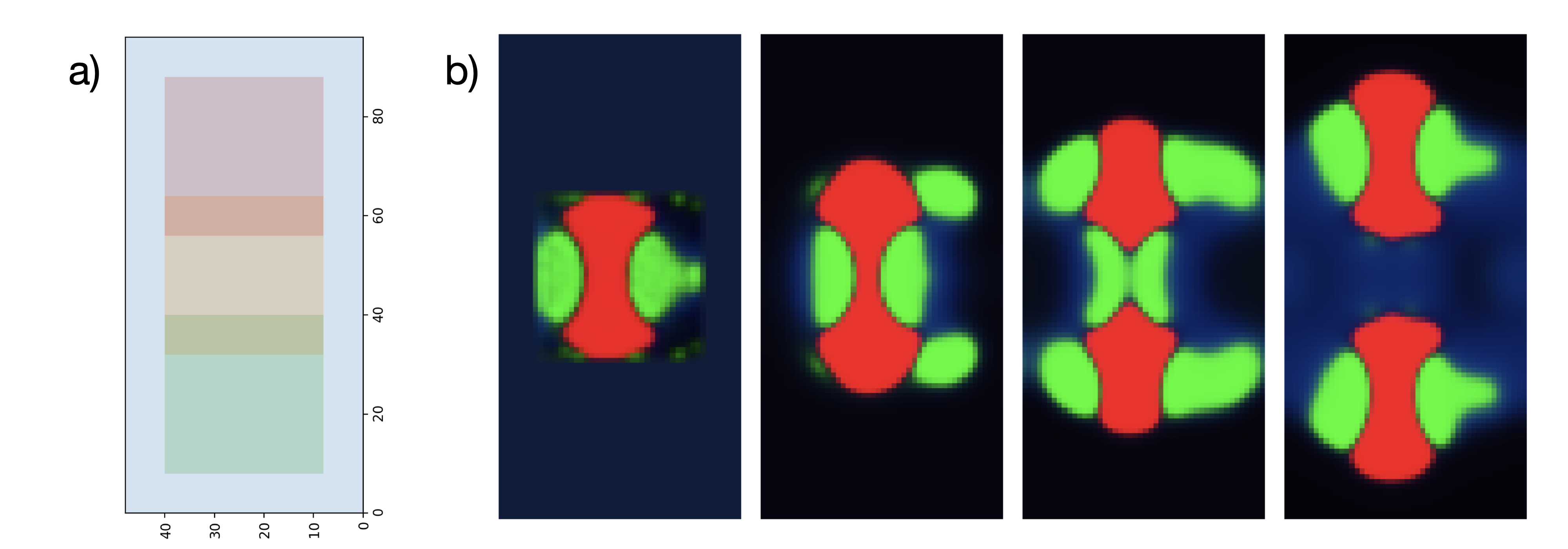}
    \caption{An example of the optimization-derived structure and associated replication-like dynamics in a 5-chemical species Dense RDN. (a) shows the structure of the loss, which requires similarity between the region in orange at $t=0$ and the two regions in red and green at $t=T$. (b) shows the resulting local concentration vs. time from $t=0s$ (left) to $t=352s$ (right). Compare the structure in the leftmost panel of b) to the two structures in the rightmost panel.}
    \label{fig:rep_fig}
\end{figure}

Thus far we have solved for specific states $\bm{X}_t$ of dynamical systems or for properties of their transitions $\Delta \bm{X}_t$. It may be interesting instead to optimize directly for properties of the full time evolution $\mathcal{X}$. Dissipative structures in reaction-diffusion models have been previously shown to undergo  particle-like motion \cite{tuckwell1979solitons}\cite{bode1995pattern}, and even `replication' of simple spot patterns \cite{pearson1993complex}\cite{lee1994experimental}\cite{virgo2011thermodynamics}. Here we seek structural reorganization and motion without explicitly specifying the dynamical model \textit{or} any of the states $\bm{X}_t$. We do this by optimizing a loss which requires similarity between $\bm{X}_0$ and \textit{two} shifted versions of $\bm{X}_T$ (Figure \ref{fig:rep_fig}a):

\begin{equation}\label{eq:rep_loss_base}
    \mathcal{L}(\vect{\theta}) = \lVert \bm{X}_0 - T_{-w}(\bm{X}_T) \rVert_2 + \left\lVert \bm{X}_0 - T_{+w}(\bm{X}_T) \right\rVert_2
\end{equation}
Where $T_{-w}(\bm{X})$ indicates a vertical translation of $\bm{X}$ by a distance $-w$. There is however a trivial solution to (\ref{eq:rep_loss_base}) which consists of uniform, time-invariant concentrations of all chemical species. We therefore introduce a loss term which encourages the spatial standard deviation in concentrations at the end of the time series to be close to or above a target value $\beta^*$:
\begin{equation}\label{eq:variance_penalty}
    \mathcal{L}_{STD}(\vect{\theta},\beta^*)=\frac{1}{N_s}\sum_i\mathrm{Max}\left(0,\left(\beta^*-\sqrt{\mathrm{Var}\left[\bm{X}_T^i\right]}\right)\right)^2
\end{equation}
Where the variance is over spatial positions $(u,v)$. We then minimize the sum of (\ref{eq:rep_loss_base}) and (\ref{eq:variance_penalty}). By requiring this similarity between a central region at the beginning of the time series and two distinct regions of $\bm{X}_T$ at the end of the time series (Figure \ref{fig:rep_fig}a) the optimization must converge to not simply translation but something vaguely akin to replication to minimize this loss. This requires optimization over longer timescales, necessitating a modified `incremental' optimization procedure (See Appendix \ref{sec:incremental_optimization}). Nonetheless the optimization converges finally to a combination of a chemical reaction network (Appendix Figure \ref{fig:crn_graphs}b) and initial conditions which roughly matches the desired dynamics (Figure \ref{fig:rep_fig}b). The replication fidelity is modest, with an average Pearson's correlation between the `parent' structure at $t=0$ and the two `daughter' structures at $t=352s$ of .893. 

\newcommand{\reppanelscale}{.25\textwidth}
\newcommand{\shftamnt}{-.9in}

This is of course \textit{not} true replication. The two `daughter' patterns on the right of Figure \ref{fig:rep_fig}b are not capable themselves of dividing, and if they were this loss structure would not handle the inevitable crowding and collisions between their offspring. Nonetheless, this serves as an example of relatively complex dynamical reorganization which emerges completely from the form of the loss and the properties of the physicochemical model.

\section{Discussion}

We have demonstrated through several examples that if a hypothesis about the possible states and dynamics of a complex system can be appropriately represented mathematically as a loss function, the search through the combinatorially vast space of possible behaviors of the system can be guided by nonconvex optimization.

This approach has a fundamental limitation; failure of the optimization to converge does not constitute falsification of the hypothesis. Rather, this is an absence of proof of the hypothesis, not a proof of absence of the hypothesized behavior.

Still, this approach allowed the optimization-driven identification of diffusion-coupled chemical reaction networks which can stably support predetermined spatial patterns (\ref{sec:pattern_formation}). While the examples we show here are not emergent but rather dependent on optimized initial conditions, flavors of our approach may nonetheless find use as improvements on previously demonstrated techniques in e.g. micro/nanofabrication \cite{epstein2016reaction}\cite{fialkowski2006principles}\cite{tikhomirov2017fractal}\cite{coli2022inverse}\cite{scalise2014designing}. Importantly, our approach does not require pre-specification of the specific reactions to include but rather allows the optimization to select them from within an initial dense reaction network.

Rather than optimizing for specific states we can also search for dynamics, and we've shown here that we can do this without specifying \textit{a priori} either the shape or motion (\ref{sec:replication}). While the form of the loss we've chosen gives ``replication-esque'' dynamics, in reality the similarity to replication is quite superficial: The replication is low-fidelity and unstable, and likely requires longer timescales, more complex CRNs, losses which specifically encourage stability, or a combination of these, in order to make the analogy to true replication less flimsy. Still, the ability to derive purely from optimization a non-isotropic concentration distribution that can dramatically re-organize in such a manner is at least a step towards demonstrating the feasibility of complex true replicators in a system without compartmentalization but rather only reaction and diffusion.

Perhaps most interestingly, this approach allows us to examine the dissipative structures that emerge when we search for models with extreme thermodynamic properties e.g. maximal entropy production rates (\ref{sec:dissipation_maximization}). Here we show only qualitative observations, and note that for emergent spatial structure in this model only the diffusion component of the dissipation rate should be maximized. These structures were found in the presence of a simple, spatiotemporally invariant drive. However it has been hypothesized that the adaptations relevant for the emergence of persistent nonequilibrium phenomena (such as life) are induced while dissipating energy from more complex, difficult-to-exploit drives \cite{perunov2016statistical}\cite{england2015dissipative}. It is possible that, if combined with appropriate spatiotemporally variable and chemically complex drives, this optimization approach could help shed light on these hypotheses.

Finally, while we focus on a specific reaction-diffusion type model here, the method presented is applicable to any differentiable model and loss function. The intersection of physics, chemistry and biology is littered with systems where reductionism has produced a simple and accurate differential model of a complex system, and yet this model has, as yet, failed to yield a comprehensive understanding  of the phenomenon it describes \cite{anderson1972more}. This knowledge gap is mirrored in an observation made by Turing himself, one of the first to investigate mathematical models of RD systems:

\begin{displayquote}
    \textit{This is the assumption that as soon as a fact is presented to a mind all consequences of that fact spring into the mind simultaneously with it. It is a very useful assumption under many circumstances, but one too easily forgets that it is false} \cite{turing1950computing}.
\end{displayquote}

In analogy, when the 'fact' of the mechanistic model of a complex system is known, and yet existing methods struggle to map the universe of dynamics that are consequences of this fact, we hope that the method described here provides a complementary approach to closing that gap in understanding.

\vspace{1cm}

All source code is available at \url{https://github.com/hunterelliott/dense-rdn}

\printbibliography

@article{pearson1993complex,
  title={Complex patterns in a simple system},
  author={Pearson, John E},
  journal={Science},
  volume={261},
  number={5118},
  pages={189--192},
  year={1993},
  publisher={American Association for the Advancement of Science}
}

@article{lee1993pattern,
  title={Pattern formation by interacting chemical fronts},
  author={Lee, Kyoung J and McCormick, WD and Ouyang, Qi and Swinney, Harry L},
  journal={Science},
  volume={261},
  number={5118},
  pages={192--194},
  year={1993},
  publisher={American Association for the Advancement of Science}
}

@article{perunov2016statistical,
  title={Statistical physics of adaptation},
  author={Perunov, Nikolay and Marsland, Robert A and England, Jeremy L},
  journal={Physical Review X},
  volume={6},
  number={2},
  pages={021036},
  year={2016},
  publisher={APS}
}

@article{england2015dissipative,
  title={Dissipative adaptation in driven self-assembly},
  author={England, Jeremy L},
  journal={Nature nanotechnology},
  volume={10},
  number={11},
  pages={919--923},
  year={2015},
  publisher={Nature Publishing Group}
}

@article{mahara2005entropy,
  title={Entropy production in a two-dimensional reversible Gray-Scott system},
  author={Mahara, Hitoshi and Yamaguchi, Tomohiko and Shimomura, Masatsugu},
  journal={Chaos: An Interdisciplinary Journal of Nonlinear Science},
  volume={15},
  number={4},
  pages={047508},
  year={2005},
  publisher={American Institute of Physics}
}

@book{kondepudi2014modern,
  title={Modern thermodynamics: from heat engines to dissipative structures},
  author={Kondepudi, Dilip and Prigogine, Ilya},
  year={2014},
  publisher={John Wiley \& Sons}
}

@article{mahara2010calculation,
  title={Calculation of the entropy balance equation in a non-equilibrium reaction-diffusion system},
  author={Mahara, Hitoshi and Yamaguchi, Tomohiko},
  journal={Entropy},
  volume={12},
  number={12},
  pages={2436--2449},
  year={2010},
  publisher={Molecular Diversity Preservation International}
}

@article{turing1952chemical,
  title={The Chemical Basis of Morphogenesis},
  author={Turing, AM},
  journal={Philosophical Transactions of the Royal Society of London. Series B, Biological Sciences},
  volume={237},
  number={641},
  pages={37--72},
  year={1952}
}

@article{prigogine1971biological,
  title={Biological order, structure and instabilities1},
  author={Prigogine, Ilya and Nicolis, Gregoire},
  journal={Quarterly reviews of biophysics},
  volume={4},
  number={2-3},
  pages={107--148},
  year={1971},
  publisher={Cambridge University Press}
}

@article{collatz1966numerical,
  title={The numerical treatment of differential equations},
  author={Collatz, Lothar},
  journal={Berlin: Springer},
  year={1966}
}

@inproceedings{kingma2015adam,
  title={Adam: A Method for Stochastic Optimization},
  author={Kingma, Diederik P and Ba, Jimmy},
  booktitle={ICLR (Poster)},
  year={2015}
}

@misc{chollet2015keras,
  title={Keras},
  author={Chollet, Fran\c{c}ois and others},
  year={2015},
  howpublished={\url{https://keras.io}},
}

@article{tuckwell1979solitons,
  title={Solitons in a reaction-diffusion system},
  author={Tuckwell, Henry C},
  journal={Science},
  volume={205},
  number={4405},
  pages={493--495},
  year={1979},
  publisher={American Association for the Advancement of Science}
}

@article{bode1995pattern,
  title={Pattern formation in reaction-diffusion systems-dissipative solitons in physical systems},
  author={Bode, Mathias and Purwins, H-G},
  journal={Physica D: Nonlinear Phenomena},
  volume={86},
  number={1-2},
  pages={53--63},
  year={1995},
  publisher={Elsevier}
}

@article{lee1994experimental,
  title={Experimental observation of self-replicating spots in a reaction--diffusion system},
  author={Lee, Kyoung-Jin and McCormick, William D and Pearson, John E and Swinney, Harry L},
  journal={Nature},
  volume={369},
  number={6477},
  pages={215--218},
  year={1994},
  publisher={Nature Publishing Group}
}

@article{szabo2002silico,
  title={In silico simulations reveal that replicators with limited dispersal evolve towards higher efficiency and fidelity},
  author={Szab{\'o}, P{\'e}ter and Scheuring, Istv{\'a}n and Cz{\'a}r{\'a}n, Tam{\'a}s and Szathm{\'a}ry, E{\"o}rs},
  journal={Nature},
  volume={420},
  number={6913},
  pages={340--343},
  year={2002},
  publisher={Nature Publishing Group}
}

@article{adamski2020self,
  title={From self-replication to replicator systems en route to de novo life},
  author={Adamski, Paul and Eleveld, Marcel and Sood, Ankush and Kun, {\'A}d{\'a}m and Szil{\'a}gyi, Andr{\'a}s and Cz{\'a}r{\'a}n, Tam{\'a}s and Szathm{\'a}ry, E{\"o}rs and Otto, Sijbren},
  journal={Nature Reviews Chemistry},
  volume={4},
  number={8},
  pages={386--403},
  year={2020},
  publisher={Nature Publishing Group}
}

@article{epstein2016reaction,
  title={Reaction--diffusion processes at the nano-and microscales},
  author={Epstein, Irving R and Xu, Bing},
  journal={Nature nanotechnology},
  volume={11},
  number={4},
  pages={312--319},
  year={2016},
  publisher={Nature Publishing Group}
}

@misc{fialkowski2006principles,
  title={Principles and implementations of dissipative (dynamic) self-assembly},
  author={Fialkowski, Marcin and Bishop, Kyle JM and Klajn, Rafal and Smoukov, Stoyan K and Campbell, Christopher J and Grzybowski, Bartosz A},
  journal={The Journal of Physical Chemistry B},
  volume={110},
  number={6},
  pages={2482--2496},
  year={2006},
  publisher={ACS Publications}
}

@article{tikhomirov2017fractal,
  title={Fractal assembly of micrometre-scale DNA origami arrays with arbitrary patterns},
  author={Tikhomirov, Grigory and Petersen, Philip and Qian, Lulu},
  journal={Nature},
  volume={552},
  number={7683},
  pages={67--71},
  year={2017},
  publisher={Nature Publishing Group}
}

@article{coli2022inverse,
  title={Inverse design of soft materials via a deep learning--based evolutionary strategy},
  author={Coli, Gabriele M and Boattini, Emanuele and Filion, Laura and Dijkstra, Marjolein},
  journal={Science advances},
  volume={8},
  number={3},
  pages={eabj6731},
  year={2022},
  publisher={American Association for the Advancement of Science}
}

@article{scalise2014designing,
  title={Designing modular reaction-diffusion programs for complex pattern formation},
  author={Scalise, Dominic and Schulman, Rebecca},
  journal={Technology},
  volume={2},
  number={01},
  pages={55--66},
  year={2014},
  publisher={World Scientific}
}

@article{kondo2010reaction,
  title={Reaction-diffusion model as a framework for understanding biological pattern formation},
  author={Kondo, Shigeru and Miura, Takashi},
  journal={science},
  volume={329},
  number={5999},
  pages={1616--1620},
  year={2010},
  publisher={American Association for the Advancement of Science}
}

@article{landge2020pattern,
  title={Pattern formation mechanisms of self-organizing reaction-diffusion systems},
  author={Landge, Amit N and Jordan, Benjamin M and Diego, Xavier and M{\"u}ller, Patrick},
  journal={Developmental biology},
  volume={460},
  number={1},
  pages={2--11},
  year={2020},
  publisher={Elsevier}
}

@article{vanag2009pattern,
  title={Pattern formation mechanisms in reaction-diffusion systems},
  author={Vanag, Vladimir K and Epstein, Irving R},
  journal={International Journal of Developmental Biology},
  volume={53},
  number={5-6},
  pages={673--681},
  year={2009},
  publisher={UPV/EHU Press}
}

@article{scheuring2003spatial,
  title={Spatial models of prebiotic evolution: soup before pizza?},
  author={Scheuring, Istv{\'a}n and Cz{\'a}r{\'a}n, Tam{\'a}s and Szab{\'o}, P{\'e}ter and K{\'a}rolyi, Gy{\"o}rgy and Toroczkai, Zolt{\'a}n},
  journal={Origins of life and evolution of the biosphere},
  volume={33},
  number={4},
  pages={319--355},
  year={2003},
  publisher={Springer}
}

@article{radford2015unsupervised,
  title={Unsupervised representation learning with deep convolutional generative adversarial networks},
  author={Radford, Alec and Metz, Luke and Chintala, Soumith},
  journal={arXiv preprint arXiv:1511.06434},
  year={2015}
}

@phdthesis{virgo2011thermodynamics,
  title={Thermodynamics and the structure of living systems},
  author={Virgo, Nathaniel D and others},
  year={2011},
  school={University of Sussex Brighton}
}

@book{feinberg2019foundations,
  title={Foundations of Chemical Reaction Network Theory},
  author={Feinberg, Martin},
  volume={202},
  year={2019},
  publisher={Springer}
}

@inproceedings{angeli2009tutorial,
  title={A tutorial on Chemical Reaction Networks dynamics},
  author={Angeli, David},
  booktitle={2009 European Control Conference (ECC)},
  pages={649--657},
  year={2009},
  organization={IEEE}
}

@article{gray1983autocatalytic,
  title={Autocatalytic reactions in the isothermal, continuous stirred tank reactor: isolas and other forms of multistability},
  author={Gray, P and Scott, SK},
  journal={Chemical Engineering Science},
  volume={38},
  number={1},
  pages={29--43},
  year={1983},
  publisher={Elsevier}
}

@article{prigogine1968symmetry,
  title={Symmetry breaking instabilities in dissipative systems. II},
  author={Prigogine, Ilya and Lefever, Ren{\'e}},
  journal={The Journal of Chemical Physics},
  volume={48},
  number={4},
  pages={1695--1700},
  year={1968},
  publisher={American Institute of Physics}
}

@article{prigogine1978time,
  title={Time, structure, and fluctuations},
  author={Prigogine, Ilya},
  journal={Science},
  volume={201},
  number={4358},
  pages={777--785},
  year={1978},
  publisher={American Association for the Advancement of Science}
}

@inproceedings{ioffe2015batch,
  title={Batch normalization: Accelerating deep network training by reducing internal covariate shift},
  author={Ioffe, Sergey and Szegedy, Christian},
  booktitle={International conference on machine learning},
  pages={448--456},
  year={2015},
  organization={PMLR}
}

@article{armstrong2004modelling,
  title={Modelling wave propagation across a series of gaps},
  author={Armstrong, Gavin R and Taylor, Annette F and Scott, Stephen K and G{\'a}sp{\'a}r, Vilmos},
  journal={Physical Chemistry Chemical Physics},
  volume={6},
  number={19},
  pages={4677--4681},
  year={2004},
  publisher={Royal Society of Chemistry}
}

@article{ertl1991oscillatory,
  title={Oscillatory kinetics and spatio-temporal self-organization in reactions at solid surfaces},
  author={Ertl, Gerhard},
  journal={Science},
  volume={254},
  number={5039},
  pages={1750--1755},
  year={1991},
  publisher={American Association for the Advancement of Science}
}

@article{hamik2003excitation,
  title={Excitation waves in reaction-diffusion media with non-monotonic dispersion relations},
  author={Hamik, Chad T and Steinbock, Oliver},
  journal={New Journal of Physics},
  volume={5},
  number={1},
  pages={58},
  year={2003},
  publisher={IOP Publishing}
}

@article{or1998spot,
  title={Spot bifurcations in three-component reaction-diffusion systems: The onset of propagation},
  author={Or-Guil, M and Bode, M and Schenk, CP and Purwins, H-G},
  journal={Physical Review E},
  volume={57},
  number={6},
  pages={6432},
  year={1998},
  publisher={APS}
}

@article{khater2002tanh,
  title={The tanh method, a simple transformation and exact analytical solutions for nonlinear reaction--diffusion equations},
  author={Khater, AH and Malfliet, W and Callebaut, DK and Kamel, ES},
  journal={Chaos, Solitons \& Fractals},
  volume={14},
  number={3},
  pages={513--522},
  year={2002},
  publisher={Elsevier}
}

@article{smith2018beyond,
  title={Beyond activator-inhibitor networks: the generalised Turing mechanism},
  author={Smith, Stephen and Dalchau, Neil},
  journal={arXiv preprint arXiv:1803.07886},
  year={2018}
}

@article{kondo2017updated,
  title={An updated kernel-based Turing model for studying the mechanisms of biological pattern formation},
  author={Kondo, Shigeru},
  journal={Journal of Theoretical Biology},
  volume={414},
  pages={120--127},
  year={2017},
  publisher={Elsevier}
}

@article{mordvintsev2021differentiable,
  title={Differentiable Programming of Reaction-Diffusion Patterns},
  author={Mordvintsev, Alexander and Randazzo, Ettore and Niklasson, Eyvind},
  journal={arXiv preprint arXiv:2107.06862},
  year={2021}
}

@inproceedings{phan2006sketching,
  title={Sketching reaction-diffusion texture},
  author={Phan, Ly and Grimm, Cindy},
  booktitle={Proceedings of the Third Eurographics conference on Sketch-Based Interfaces and Modeling},
  pages={107--114},
  year={2006}
}

@article{mordvintsev2021texture,
  title={Texture Generation with Neural Cellular Automata},
  author={Mordvintsev, Alexander and Niklasson, Eyvind and Randazzo, Ettore},
  journal={arXiv preprint arXiv:2105.07299},
  year={2021}
}

@article{bartocci2016formal,
  title={A formal methods approach to pattern recognition and synthesis in reaction diffusion networks},
  author={Bartocci, Ezio and Gol, Ebru Aydin and Haghighi, Iman and Belta, Calin},
  journal={IEEE Transactions on Control of Network Systems},
  volume={5},
  number={1},
  pages={308--320},
  year={2016},
  publisher={IEEE}
}

@article{scalise2016emulating,
  title={Emulating cellular automata in chemical reaction--diffusion networks},
  author={Scalise, Dominic and Schulman, Rebecca},
  journal={Natural Computing},
  volume={15},
  number={2},
  pages={197--214},
  year={2016},
  publisher={Springer}
}

@article{scholes2017three,
  title={A three-step framework for programming pattern formation},
  author={Scholes, Natalie S and Isalan, Mark},
  journal={Current opinion in chemical biology},
  volume={40},
  pages={1--7},
  year={2017},
  publisher={Elsevier}
}

@article{murphy2018synthesizing,
  title={Synthesizing and tuning stochastic chemical reaction networks with specified behaviours},
  author={Murphy, Niall and Petersen, Rasmus and Phillips, Andrew and Yordanov, Boyan and Dalchau, Neil},
  journal={Journal of The Royal Society Interface},
  volume={15},
  number={145},
  pages={20180283},
  year={2018},
  publisher={The Royal Society}
}

@article{chua1995autonomous,
  title={Autonomous cellular neural networks: a unified paradigm for pattern formation and active wave propagation},
  author={Chua, Leon O and Hasler, Martin and Moschytz, George S and Neirynck, Jacques},
  journal={IEEE Transactions on Circuits and Systems I: Fundamental theory and applications},
  volume={42},
  number={10},
  pages={559--577},
  year={1995},
  publisher={IEEE}
}

@article{li2020reaction,
  title={Reaction diffusion system prediction based on convolutional neural network},
  author={Li, Angran and Chen, Ruijia and Farimani, Amir Barati and Zhang, Yongjie Jessica},
  journal={Scientific reports},
  volume={10},
  number={1},
  pages={1--9},
  year={2020},
  publisher={Nature Publishing Group}
}

@article{pathak2018model,
  title={Model-free prediction of large spatiotemporally chaotic systems from data: A reservoir computing approach},
  author={Pathak, Jaideep and Hunt, Brian and Girvan, Michelle and Lu, Zhixin and Ott, Edward},
  journal={Physical review letters},
  volume={120},
  number={2},
  pages={024102},
  year={2018},
  publisher={APS}
}

@article{pathak2017using,
  title={Using machine learning to replicate chaotic attractors and calculate Lyapunov exponents from data},
  author={Pathak, Jaideep and Lu, Zhixin and Hunt, Brian R and Girvan, Michelle and Ott, Edward},
  journal={Chaos: An Interdisciplinary Journal of Nonlinear Science},
  volume={27},
  number={12},
  pages={121102},
  year={2017},
  publisher={AIP Publishing LLC}
}

@inproceedings{zakeri2019weakly,
  title={Weakly supervised learning technique for solving partial differential equations; case study of 1-d reaction-diffusion equation},
  author={Zakeri, Behzad and Monsefi, Amin Karimi and Samsam, Sanaz and Monsefi, Bahareh Karimi},
  booktitle={International Congress on High-Performance Computing and Big Data Analysis},
  pages={367--377},
  year={2019},
  organization={Springer}
}

@article{yeo2019deep,
  title={Deep learning algorithm for data-driven simulation of noisy dynamical system},
  author={Yeo, Kyongmin and Melnyk, Igor},
  journal={Journal of Computational Physics},
  volume={376},
  pages={1212--1231},
  year={2019},
  publisher={Elsevier}
}

@article{dauphin2014identifying,
  title={Identifying and attacking the saddle point problem in high-dimensional non-convex optimization},
  author={Dauphin, Yann N and Pascanu, Razvan and Gulcehre, Caglar and Cho, Kyunghyun and Ganguli, Surya and Bengio, Yoshua},
  journal={Advances in neural information processing systems},
  volume={27},
  year={2014}
}

@article{turing1950computing,
    author = {Turing, A. M.},
    title = "{I.—COMPUTING MACHINERY AND INTELLIGENCE}",
    journal = {Mind},
    volume = {LIX},
    number = {236},
    pages = {433-460},
    year = {1950},
    publisher={Oxford Academic}
}

@article{anderson1972more,
  title={More is different: broken symmetry and the nature of the hierarchical structure of science.},
  author={Anderson, Philip W},
  journal={Science},
  volume={177},
  number={4047},
  pages={393--396},
  year={1972},
  publisher={American Association for the Advancement of Science}
}

@article{volpert2009reaction,
  title={Reaction--diffusion waves in biology},
  author={Volpert, Vitaly and Petrovskii, Sergei},
  journal={Physics of life reviews},
  volume={6},
  number={4},
  pages={267--310},
  year={2009},
  publisher={Elsevier}
}

@article{simoyi1982one,
  title={One-dimensional dynamics in a multicomponent chemical reaction},
  author={Simoyi, Reuben H and Wolf, Alan and Swinney, Harry L},
  journal={Physical Review Letters},
  volume={49},
  number={4},
  pages={245},
  year={1982},
  publisher={APS}
}

@article{rossler1976chemical,
  title={Chemical turbulence: chaos in a simple reaction-diffusion system},
  author={R{\"o}ssler, Otto E},
  journal={Zeitschrift f{\"u}r Naturforschung A},
  volume={31},
  number={10},
  pages={1168--1172},
  year={1976},
  publisher={De Gruyter}
}

\appendix

\appendixpage

\section{Miscellaneous Methods Details}

Here we provide further verbal and mathematical description of key aspects of the methods. Additional detail can be found in the source code, available at \url{https://github.com/hunterelliott/dense-rdn}.

\subsection{Minimizing Numerical Integration Error}\label{sec:step_size}

With highly nonlinear differential models and a simple Euler integration scheme, we must be careful to ensure that the results of the optimizations are still reflective of the physicochemical model and not simply artifacts of numerical integration error. We minimize the effect of these errors in several ways.

First, we control the step size $\Delta \bm{X}_t$ via penalty terms in the loss. In \ref{sec:pattern_formation} this is a natural part of the loss (Eq. \ref{eq:target_stable_loss}) which encourages stable pattern formation, and we set $\lambda_\Delta=100$ in those experiments. For the remaining experiments we introduce a penalty only for step sizes that exceed a threshold $\delta_{Max}$:

\begin{equation}\label{eq:dxdt_penalty}
    \mathcal{L}_{\delta}(\vect{\theta},\delta_{Max})=\lambda_\delta \frac{1}{UVT} \sum_{(u,v), t} \mathrm{Max}\left[ \left| \Delta\bm{X}_t (u,v)\right| - \delta_{Max}, 0 \right] 
\end{equation}

Where we used $\lambda_\delta=1000$ and $\delta_{Max}=.05$ for all experiments except those in \ref{sec:replication} which used $\lambda_\delta=1.0$ and $\delta_{Max}=.3$. The full loss then is the sum of $\mathcal{L}_{\delta}$ and the loss described for each experiment.

Additionally, for all results presented we verified that the results were unchanged if we ran forward simulations with the time step $\Delta t$ decreased by two orders of magnitude (from 1.0 to .01). This decreased time step should dramatically reduce numerical error and so the fact that our results are unchanged indicates they are unlikely to be artifactual.

\subsection{Local Concentration Visualizations}\label{sec:conc_vis}

When the number of chemical species to be visualized $N_s$ is $>$3 we first perform a principal component analysis (PCA) projection of the concentration dimensions. We retain only the largest  3 components and map these to the intensities of the red, green and blue channels, respectively, of an RGB image for visualization. In all cases presented here these three components explained $>$ 90\% of the concentration variance. Note that this approach allows approximate visualization of arbitrarily complex chemical species mixtures, but precludes inclusion of a concentration color scale bar. This is however unimportant, given the fact that both the concentrations and reaction rate constants are determined simultaneously in the optimization; they could both be arbitrarily re-scaled and so we are effectively working in arbitrary concentration units.

\subsection{Diffusion Modeling}\label{sec:diffusion_modeling}

We used an isotropic discrete approximation of the Laplacian operator \cite{collatz1966numerical} to approximate Eq. (\ref{eq:diff_update}). In all cases diffusion coefficients were constrained to lie between $.05$ and $.02\times10^{-5} \mathrm{m^2/s}$. The lower bound serves to prevent runaway accumulation of chemical species and spatial discontinuities in concentration. The upper bound was chosen to prevent numerical artifacts (checkerboard patterns, oscillation) given the $1\mathrm{s}$ time step and discrete Laplacian approximation. Using these units for the diffusion coefficients the physical dimension of a single spatial element of $\bm{X}$ (the `pixel size') is $.01\mathrm{m}$, but given that the diffusion coefficients and concentrations are all optimized simultaneously and could be arbitrarily re-scaled, all the units are effectively arbitrary. All spatial domains were $64\times64$ pixels except for in \ref{sec:replication} where the dimensions were $48\times96$.

\subsection{Neural Networks}\label{sec:dnns}

The generator model $G(\vect{z},\vect{\theta}_G)$ is a simple architecture consisting of repeated blocks of stride 2 convolution transpose, batch normalization \cite{ioffe2015batch} and a $tanh$ activation. The final generated spatial domain is therefore $2^{N_{blocks}}$ in both width and height, and the number of blocks varies within the presented results accordingly. The output is also filtered with a fixed 2D gaussian kernel ($\sigma=1.0 $ pixels), to avoid introducing spatial discontinuities and therefore numerical error in diffusion simulations and diffusion entropy rate calculations. Finally, we exclude batch norm from the last layer (as in DCGAN \cite{radford2015unsupervised}) and instead add only an optimizable scaling that is constrained to prevent negative concentrations and limit the maximum generated concentration to 10.0 (in arbitrary units).

The encoder models used in \ref{sec:inf_transmit} use a similar structure, with stride 2 convolutions followed by a $tanh$ activation and batch normalization. 

Experimentation was not extensive and constrained by hardware limitations, but in general we saw no obvious improvement in convergence properties or final optimized values from using ReLU activations or higher capacity generator or encoder architectures.

\section{Chemical Reaction Networks}\label{sec:crn_methods}

\subsection{Introductory Example Reaction Network}\label{sec:complex_example_description}

The dynamics shown in Figure \ref{fig:complex_example} are derived from a `coupled Gray-Scott' reaction system, which consists of two standard Gray-Scott reaction systems \cite{gray1983autocatalytic} linked by an additional reaction which allows interconversion of the two autocatalysts.

\subsection{Dense Chemical Reaction Networks}\label{sec:dense_crns}

The reaction networks used here are ``dense'' in the sense that they contain all possible reactions matching a particular prototype. That is, if a reaction network contained species $\mathcal{S} = \{A,B,C,D\}$ then for the first reaction in \eqref{eq:crn_sketch} we would include:

\begin{equation}
    \begin{aligned}
        A + B &\rightleftarrows 2C \\   
        A + C &\rightleftarrows 2B \\
        B + C &\rightleftarrows 2A \\
        A + B &\rightleftarrows 2D \\
        &...
    \end{aligned}
\end{equation}

While for the second reaction $A\rightleftarrows B$ we would include every possible reversible unimolecular interconversion e.g. $A\rightleftarrows B$, $A\rightleftarrows C$, $A\rightleftarrows D$, $B\rightleftarrows C$, and so on. More formally, we would generate reaction sets matching each of the reaction prototypes given in \eqref{eq:crn_sketch}:

\begin{equation}
    \begin{aligned}
        \mathcal{R}_{3.1} &= \{ X + Y \rightleftarrows 2Z | \forall X,Y,Z \in \mathcal{S}, X \ne Y, Y \ne Z, X \ne Z \} \\
        \mathcal{R}_{3.2} &= \{ X \rightleftarrows Y | \forall X,Y \in \mathcal{S}, X \ne Y \} \\
        \mathcal{R}_{3.3} &= \{ X + 2Y \rightleftarrows 3Y | \forall X,Y \in \mathcal{S}, X \ne Y \} \\
    \end{aligned}
\end{equation}
Every reaction in these sets is governed by independent forward and reverse reaction rates which are free parameters determined during the optimization. This gives e.g. a total of 40 reactions for a 5-species dense CRN (including forward and reverse), and 24 reactions for a 4 species dense CRN.

To model the dynamics of these chemical reactions we assume standard mass-action kinetics. Using the matrix representations commonly used in chemical reaction network theory, the reaction rate calculations $\Delta\bm{X}_{t_{Rxn}}$ are matrix multiplications \cite{feinberg2019foundations}\cite{angeli2009tutorial},  which are efficiently calculated on the GPUs used for both forward simulation and optimization.

\subsection{CRN Motivation}\label{sec:crn_motivation}

We chose the reaction network described above both for the variety of reactions which comprise it, as well as it's potential for diverse behavior. This gives the optimization process a broad space of possible reaction sets and dynamics to explore, allowing for the possibility of interesting behavior without pre-specifying it.

The reactions include many if not most stereotypical uni- bi- and tri-molecular reactions. Given sufficient timescale, more complex reaction mechanisms can be approximated with these more elementary reactions. Of more specific interest they include, as a subset, the reactions from well-studied reaction-diffusion models exhibiting complex behavior such as the Gray-Scott reaction system \cite{gray1983autocatalytic} or the Brussellator \cite{prigogine1968symmetry}. It is therefore not surprising that chemical reaction network theory tells us, on the basis of the topology and stoichiometry of these reactions, that complex dynamics are at least possible \cite{feinberg2019foundations}.

\subsection{Optimized Reaction Networks}\label{sec:crn_graphs}

The optimization process effectively chooses a specific reaction network from the many possible networks the model described above can represent. We can visualize this reaction network as a graph, with arrows pointing from reactants to products, and with the style of arrow indicating the `strength' of that reaction: Darker, larger arrows indicate larger reaction rate constants, and the graph is laid out such that strongly reacting species should be closer together (shorter arrows).

\begin{figure}[h]
    \centering
    \includegraphics[width=\textwidth]{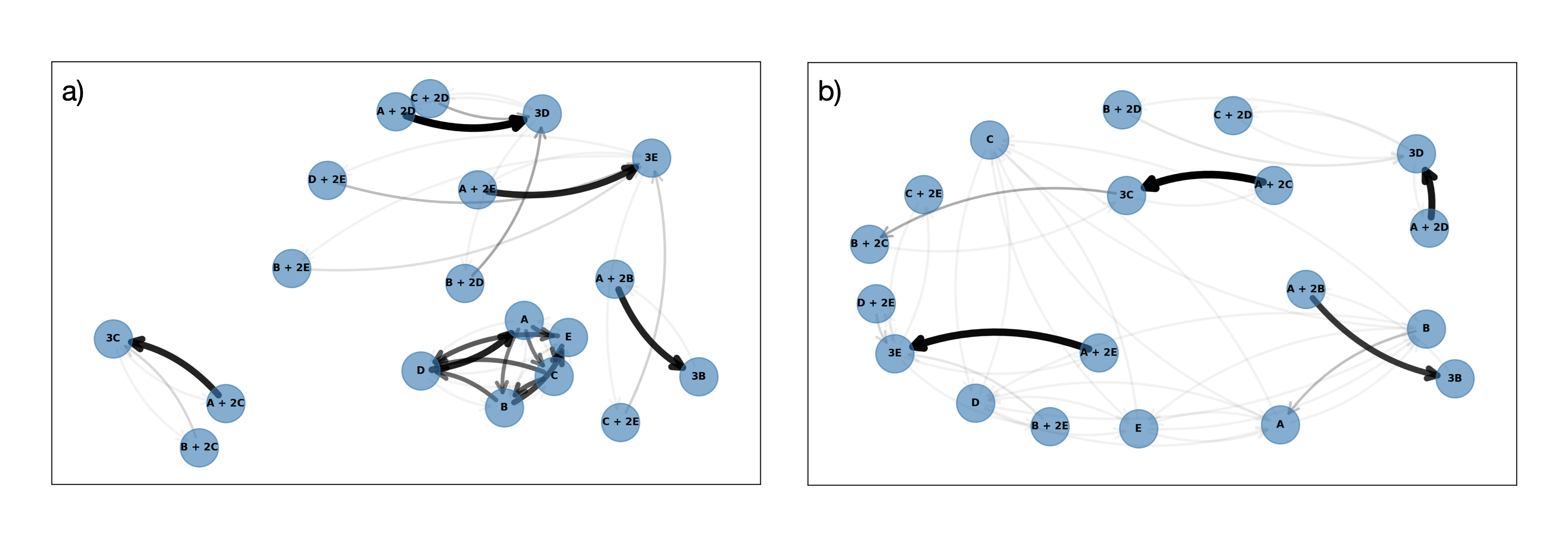}
    \caption{Visualization of the optimized reaction network graphs from dissipation maximization (a) corresponding to the results in Figure \ref{fig:diss_max}a as well as the reaction network from replication-like dynamics (b) corresponding to the results in Figure \ref{fig:rep_fig}. Darker, thicker arrows indicate larger reaction rate constants.}
     \label{fig:crn_graphs}
\end{figure}

This allows us to compare the optimized reaction network from, for example, the dissipation-maximizing loss used in \ref{sec:dissipation_maximization}  (Figure \ref{fig:crn_graphs}a), to that from the replication-like-dynamics-inducing loss used in \ref{sec:replication} (Figure \ref{fig:crn_graphs}b). It is worth noting that these are just the reaction rate constants, and the net flux through these reaction pathways depends on the state $\bm{X}$ and its time progression.

\section{Thermodynamics}\label{sec:therm}

The instantaneous entropy production rate of a single volume element $\bm{X}(u,v)$ due to chemical reactions is given by \cite{kondepudi2014modern}\cite{mahara2010calculation}:
\begin{equation}\label{eq:rxn_entropy_rate}
    \left(\frac{\partial S}{\partial t}\right)_{Rxn} = \sigma_{Rxn}=k_B\sum_{j} \left(\nu_j^+ - \nu_j^-\right)\ln\left(\frac{\nu_j^+}{\nu_j^-}\right) 
\end{equation}
Where $\nu_j^+$ and $\nu_j^-$ are the local forward and reverse reaction rates of reaction $j$ in that volume element and and $k_B$ is Boltzmann's constant, giving $\frac{\partial S}{\partial t}$ units of $\frac{\mathrm{J}}{\mathrm{K}\cdot \mathrm{L} \cdot \mathrm{s}}$. In practice we replace $\nu_j$ with $\mathrm{Max}\left(\nu_j,1\times 10^{-5}\right)$ for numerical stability.

The instantaneous local entropy production rate due to diffusion in a volume element $\bm{X}(u,v)$ is \cite{mahara2010calculation}:
\begin{equation}\label{eq:dif_entropy_rate}
    \left(\frac{\partial S}{\partial t}\right)_{Dif} = \sigma_{Dif}=k_B\sum_{i} \frac{D_i}{\bm{X}^i(u,v)}\left(\nabla \bm{X}^i(u,v) \right)^2
\end{equation}
Where again $D_i$ is the diffusion coefficient for species $i$ and $\bm{X}^i(u,v)$ is the concentration of species $i$ at position $(u,v)$. We add a small numerical stabilizer (0.1) to the denominator of (\ref{eq:dif_entropy_rate}) to avoid division by zero.

The total entropy production rate is simply the sum of these two components:
\begin{equation}
    \sigma_{Tot} = \sigma_{Rxn} + \sigma_{Dif} 
\end{equation}
Rather than integrate this rate to produce a net change in entropy $\Delta S_{Tot} = \sigma_{Tot} \Delta t$ we report and maximize the average of the instantaneous rates to avoid explicit dependence on the time and length scale of optimization.

\section{Optimization}\label{sec:optimization}

\subsection{Initialization}\label{sec:initialization}

As is commonly observed in neural network optimization, we found that initialization was important for convergence. Specifically, in our case models initialized with uniform random but low reaction rates ($k_{fi}, k_{ri}<1\times10^{-3} \hspace{.2cm} \forall \hspace{.2cm}i)$ and similarly low drive flow rates and concentrations failed to converge. Models initialized to more highly dissipative regimes - with higher reaction rates and higher drive flow rates - converged more reliably. Exhaustive exploration of initialization dependence is prohibitively computationally intensive, so we instead mimicked the kinetic parameters used in \cite{pearson1993complex}, generalizing it to our dense CRNs: One autocatalytic reaction per chemical species was initialized with a rate constant of 1. All other reaction rates were set to $1\times10^{-3} + \epsilon_r$ with $\epsilon_r \sim \mathrm{Unif}\left(-1\times10^{-4},1\times10^{-4}\right)$ as this was found to be low enough to prevent large $\Delta\bm{X}$ and the associated numerical inaccuracies and/or runaway reaction rates. The small uniform random noise $\epsilon_r$ breaks symmetry in the reaction rates. 

Feed concentrations $x_i$ for the flow-reactor drive were initialized to $\frac{1}{4^{N_{a_i}}}$ where $N_{a_i}$ is the number of autocatalytic reactions \textit{producing} species $i$. This scaling was found to produce a highly dissipative yet somewhat kinetically and numerically stable initial state.

Diffusion coefficients were initialized from $\mathrm{Unif}\left(.05,.2\right)\times 10^{-5} \mathrm{m^2/s}$, uniformly within the numerically stable regime.

\subsection{Heuristics and Hyperparameters}\label{sec:hyperparam}

\begin{table}[h]
    \begin{center}
    \begin{tabular}{c|c|c|c|c|c|c}
        \hline
        Figure & $T$ & lr & lr decrease by $\frac{1}{2}$ at & lr patience & es patience & ADAM $\beta_1$ \\
        \hline
        \ref{fig:fixed_ilya}a & 32 & $1\times 10^{-3}$ & $(5, 10, 15)\times 10^2$ & $5\times 10^{2}$ & $1\times 10^{3}$ & .95 \\
        \ref{fig:fixed_ilya}b & 64 & $1\times 10^{-3}$ & $(5, 10, 15)\times 10^2$ & $5\times 10^{2}$ & $1\times 10^{3}$ & .95 \\
        \ref{fig:diss_max}a & 64 & $1\times 10^{-3}$ & - & $2\times 10^{3}$ & $6\times 10^{3}$ & .995 \\
        \ref{fig:diss_max}b & 64 & $1\times 10^{-3}$ & - & $2\times 10^{3}$ & $6\times 10^{3}$ & .995 \\
        \ref{fig:diss_dist} & 128 & $2.5\times 10^{-4}$ & $(1, 2)\times 10^4$ & $2\times 10^{3}$ & $6\times 10^{3}$ & .995 \\
        \hline
    \end{tabular}
    \caption{Hyperparameters for presented results. lr: learning rate, lr decrease: iterations at which learning rate was decreased on a fixed schedule, lr patience: patience parameter for automatic learning rate decreases, es patience: patience parameter for automatic early stopping. lr decrease and patience columns are in units of iterations ('epoch' has no meaning in this context).}
    \label{tab:hyperparameters}
    \end{center}
\end{table}
We used the ADAM optimizer \cite{kingma2015adam}, with learning rates and moving averages $\beta_1$ as given in Table \ref{tab:hyperparameters} ($\beta_2$ of .999 was used in all experiments). With these highly nonlinear dynamical models we found that, especially for optimizations with longer timescales, gradient clipping was essential for stability of convergence. We used gradient norm clipping at 0.5 as implemented in Keras \cite{chollet2015keras}. Learning rates were automatically decreased and optimization was automatically terminated via Keras' callbacks with patience parameters as given in Table \ref{tab:hyperparameters}. In some experiments we used RMSD instead of the L2 norm given in Eq. \ref{eq:target_loss} to simplify changing the size of the spatial domain without altering the magnitude of the loss, and found this gave similar results to the L2 norm as expected.

Calculating gradients with these models requires backpropagation through the entire time series $\mathcal{X}$, introducing significant memory overhead. For this reason, and because in most cases we are optimizing for a single $\bm{X_0}$, we set the batch size to 1 except for in Sect. \ref{sec:inf_transmit}, where we used 2. For experiments where dissipation losses were included, $\lambda_\sigma$ was set to 1.0 except for in \ref{sec:inf_transmit} where it was 0.04.
During optimization the reaction rate constants and feed concentrations were constrained to be non-negative, and the flow rate was constrained to between .01 and 1.0. Diffusion coefficients were constrained to within their initialization range (\ref{sec:diffusion_modeling}).

\subsection{Incremental Optimization}\label{sec:incremental_optimization}
\begin{table}[h]
    \begin{center}
    \begin{tabular}{c|c|c|c|c}
        \hline
        Optimization & $T$ & $\beta^*$ & lr & lr decrease by $\frac{1}{2}$ at \\
        \hline
        1 & 256 & 1.0 & $1\times 10^{-3}$ & $(1, 2, 3, 50, 100, 150)\times 10^2$ \\
        2 & 320 & 1.0 & $5\times 10^{-6}$ & $(5, 10)\times 10^2$ \\
        3 & 352 & 0.5 & $1\times 10^{-6}$ & $(20, 40)\times 10^2$ \\
        \hline
    \end{tabular}
    \caption{The incremental optimization stages that produced the results in Figure \ref{fig:rep_fig}. The result from each optimization was used to initialize the next.}
    \label{tab:incremental_optimization}
    \end{center}
\end{table}
For the experiments shown in \ref{sec:replication} we must optimize over timescales long enough for the full replication-like dynamics to occur. Optimization from a random initialization above $\sim250$ time points (equivalent to 250 seconds) was highly unstable, in large part due to an inability to balance vanishing and exploding gradients in the highly nonlinear CRN. Convergence was finally achieved via an ``incremental'' optimization approach, where we first optimize at shorter timescales and then use the parameters \vect{\theta^*} from shorter timescales to initialize optimization at longer timescales. Additionally, at the longest timescales the variance target $\beta^*$ seemed to be at odds with the replication fidelity loss (Eq. \ref{eq:rep_loss_base}), and decreasing this value gave better parent-daughter correlations at convergence. It is worth noting that the large number of iterations required resulted in long optimization; the final result presented here required $>600\mathrm{k}$ weight updates in total for a wall clock time of more than 26 days. The hyperparameters used for each optimization that produced Figure \ref{fig:rep_fig} are given in Table \ref{tab:incremental_optimization}. The ADAM optimizer $\beta_1$ was set to .95 for all optimizations. 

\subsection{Control Analyses}\label{sec:controls}

We performed additional analyses to test the necessity and sufficiency of the optimization of each component of the models. These computational ``ablation experiments'' correspond to several conditions:

\begin{itemize}
    \item ``Optimized'' - All components optimized.
    \item ``Random $\bm{X_0}$'' - Optimized initial conditions $\bm{X_0}$ replaced with normally distributed random noise with the same mean and variance as the optimized initial conditions (truncated to prevent negative concentrations).
    \item ``Initial $\vect{\theta}_\Psi$'' - Optimized reaction network parameters and diffusion coefficients replaced with their values from the start of optimization (as described in \ref{sec:initialization}).
    \item ``Random $\vect{\theta}_\Psi$'' - Optimized reaction network parameters and diffusion coefficients replaced with normally distributed random noise with the same mean and variance as the optimized parameters (truncated to prevent negative reaction rates or diffusion coefficients outside the numerically stable range).
\end{itemize}

\begin{figure}[h]
    \centering
    \includegraphics[width=\textwidth]{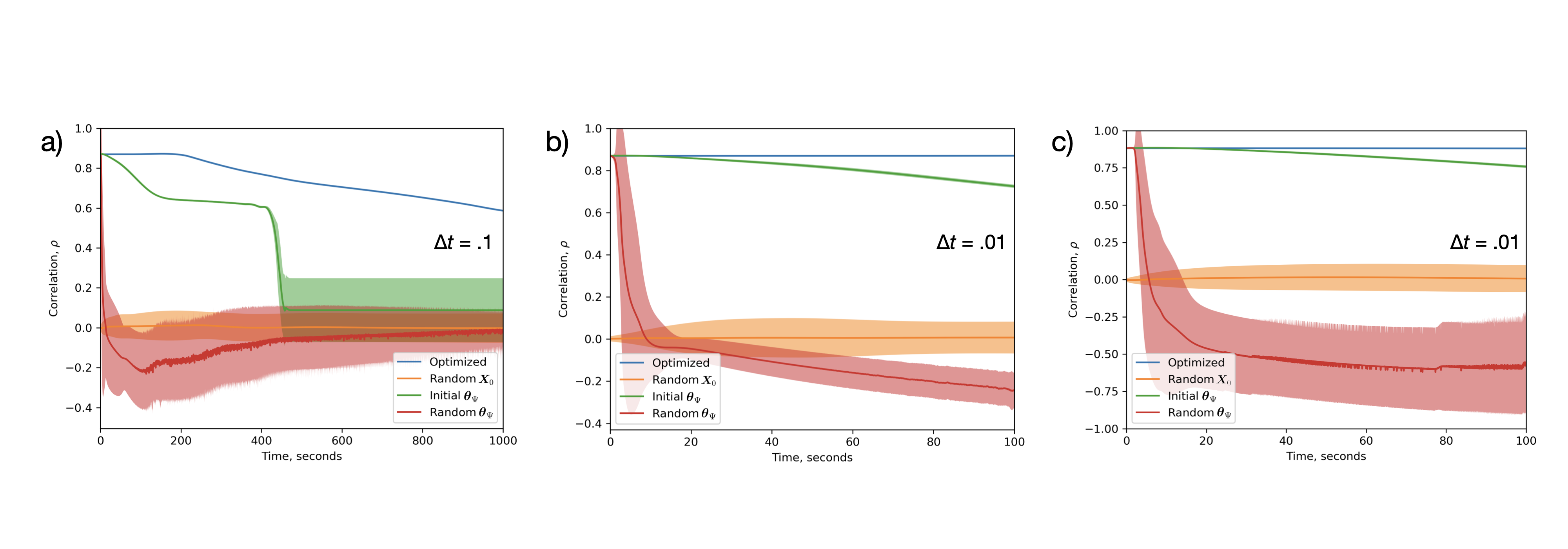}
    \caption{(a) Correlation with target vs time for the example in Figure \ref{fig:fixed_ilya}b. Both this panel and Figure \ref{fig:fixed_ilya}c use one tenth of the $1s$ time step used during optimization. (b-c) Further decreasing the time step to $.01s$ for the model from Figure \ref{fig:fixed_ilya}b (shown in b) and for Figure \ref{fig:fixed_ilya}a (shown in c)  provides additional evidence the differences are not due to numerical error. Solid lines show means, bands show +/- one standard deviation over $n=32$ randomizations.}
     \label{fig:fixed_ilya_appendix}
\end{figure}

The results of these experiments are shown in Figure \ref{fig:fixed_ilya}c and Figure \ref{fig:diss_max}c with some additional shown here and described below. These results confirm that all optimized components are required for persistent correlation with the target and for persistently high dissipation rates.
\begin{figure}[b!]
    \centering
    \includegraphics[width=\textwidth]{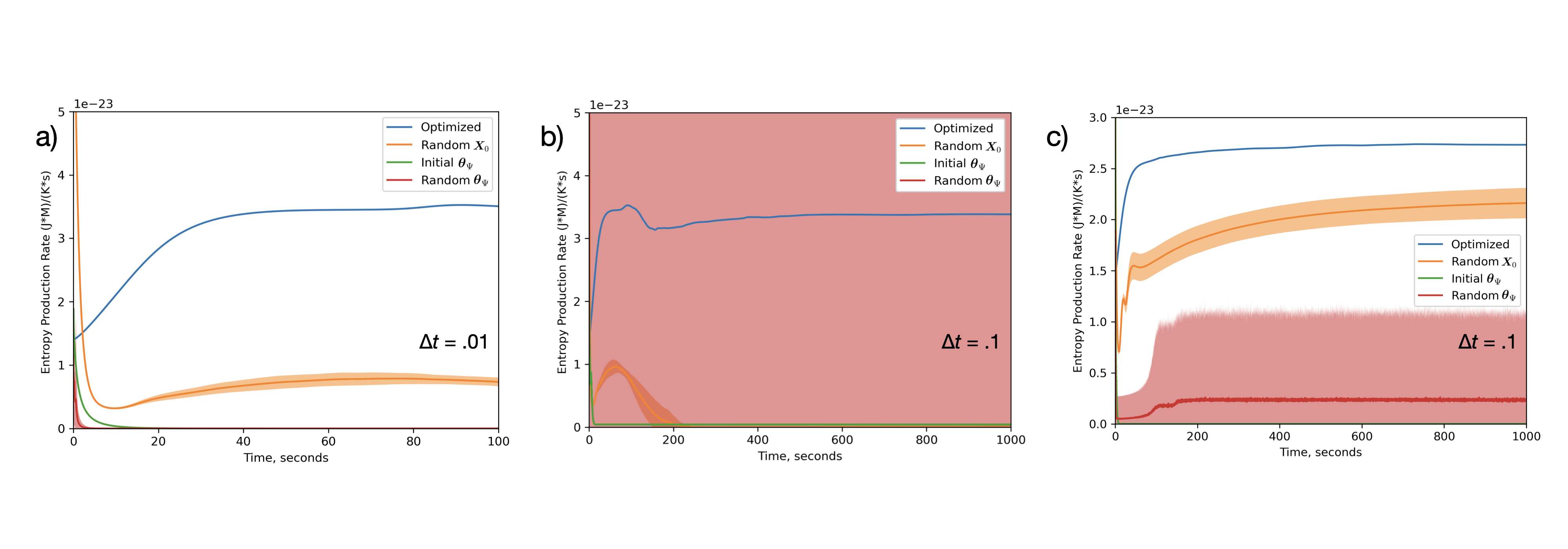}
    \caption{(a) Dissipation rate vs. time for the example in Figure \ref{fig:diss_max}b. Both this panel and \ref{fig:diss_max}c use $1/100th$ the $1s$ time step used during optimization, to confirm the differences are not due to numerical error. (b-c) Increasing the timescale by more than a factor of 10 confirms that these are stable nonequilibrium states and that the increase in dissipation rate provided by the optimization is persistent for the models from both Figure \ref{fig:diss_max}b (shown in b) and Figure \ref{fig:diss_max}a (shown in c). Solid lines show means, bands show +/- one standard deviation over $n=32$ randomizations. Note that with the $.1s$ time step in b) and c) the models with random kinetics show high variability, often due to numerical error (broad red bands).}
     \label{fig:diss_max_appendix}
\end{figure}

In Figure \ref{fig:fixed_ilya_appendix}a we show the control experiments as in the main text but for the second example (from Figure \ref{fig:fixed_ilya}b). We then repeat the same analyses for both examples but with the time step now two orders of magnitude shorter, to further reinforce that the stability and fidelity is not a result of the optimization exploiting numerical error (Figure \ref{fig:fixed_ilya_appendix}b-c). 

In Figure \ref{fig:diss_max_appendix}a we show the same result as the main text but for the second example (from Figure \ref{fig:diss_max}b). We also run both examples at a longer timescale to demonstrate that the optimized states are stably dissipative and produce entropy at consistently higher rates than those with non-optimized components (Figure \ref{fig:diss_max_appendix}b-c). Note that in this case the models with mean and standard deviation-matched random kinetic parameters (``Random $\vect{\theta}_\Psi$'', shown in red) are highly unstable and display significant numerical integration error (as evidenced by their behavior changing with a smaller time step). This further reinforces both the necessity and effectiveness of the numerical error avoidance methods described in Section \ref{sec:step_size}.

\end{document}